\documentclass{article}

\usepackage{ijcai26}

\usepackage{times}
\usepackage{soul}
\usepackage{url}
\usepackage[hidelinks]{hyperref}
\usepackage[utf8]{inputenc}
\usepackage[small]{caption}
\usepackage{graphicx}
\usepackage{amsmath}
\usepackage{amsthm}
\usepackage{booktabs}
\usepackage{algorithm}
\usepackage{algorithmic}
\usepackage[switch]{lineno}

\usepackage[utf8]{inputenc} 
\usepackage[T1]{fontenc}    
\usepackage{url}            
\usepackage{booktabs}       
\usepackage{amsfonts}       
\usepackage{nicefrac}       
\usepackage{microtype}      
\usepackage{xcolor}         
\usepackage{graphicx}
\usepackage{bm}
\usepackage{subfig}
\usepackage{booktabs}
\usepackage{makecell}
\usepackage{multicol}
\usepackage{multirow}
\usepackage{wrapfig}
\usepackage{graphicx} 
\usepackage{subfiles}
\usepackage{threeparttable} 
\newcommand{\methodname}{scLLM-DSC}

\title{scLLM-DSC: LLM-Knowledge Enhanced Cross-Modal Deep Structural Clustering \\for Single-Cell RNA Sequencing}

\author{
Ping Xu$^{1,2}$\and
Pengjiang Li$^{1,2}$\and
Tian Du$^{1,2}$\and
Zaitian Wang$^{1,2}$\and
Jiawei Gu$^4$\and
Zhiyuan Ning$^5$\and \\
Ziyue Qiao$^4$\and
Pengfei Wang$^{1,2,3}$\And
Yuanchun Zhou$^{1,2,3}$\footnotemark[1]\\
\affiliations
$^1$Computer Network Information Center, Chinese Academy of Sciences, Beijing, China\\
$^2$University of Chinese Academy of Sciences, Beijing, China\\
$^3$Hangzhou Institute for Advanced Study, University of Chinese Academy of Sciences, Hangzhou, China\\
$^4$School of Computing and Information Technology, Great Bay University, Dongguan, China\\
$^5$School of Engineering, Westlake University, Hangzhou, China
\emails
xuping0098@gmail.com, zyc@cnic.cn
}

\begin{document}

\maketitle
\setcounter{footnote}{0}
\renewcommand{\thefootnote}
{\fnsymbol{footnote}}
\footnotetext[1]{Corresponding authors.}

\begin{abstract}
Clustering is fundamental to scRNA-seq analysis, serving as a cornerstone for identifying cell populations and resolving tissue heterogeneity.
However, existing methods focus on mining numerical statistical patterns, suffering from \textit{semantic agnosticism} by neglecting the intrinsic biological functions encoded by genes. While Large Language Models (LLMs) offer promising semantic capabilities, their direct adaptation to cell clustering is hindered by the structural mismatch between generative pre-training objectives and discriminative downstream tasks. 
To bridge this gap, we propose~\textbf{\methodname}, a novel \textbf{LLM}-Knowledge Enhanced Cross-Modal \textbf{D}eep \textbf{S}tructural \textbf{C}lustering framework. 
Diverging from data-driven paradigms,~\methodname~establishes a semantically-grounded representation by synergizing two views: a \textit{Knowledge-Driven Semantic View} derived from NCBI gene priors and contextualized Cell2Sentence embeddings, and a \textit{Structure-Aware Topological View} extracted via a graph-guided encoder. Crucially, we introduce a cross-modal contrastive alignment mechanism to enforce consistency between biological semantics and transcriptomic features within a unified latent space. 
Extensive benchmarks demonstrate that~\methodname~significantly outperforms eleven state-of-the-art baselines in clustering accuracy. 
\end{abstract}


\section{Introduction}
Single-cell RNA sequencing (scRNA-seq) has revolutionized the characterization of tissue heterogeneity by providing high-resolution cellular atlases~\cite{svensson2018exponential}. 
As a pivotal analytical step, cell clustering aims to define functional identities based on expression similarity, serving as the cornerstone for downstream biological discovery~\cite{kiselev2019challenges}.

In retrospect, the evolution of clustering algorithms has been a pursuit of optimal numerical representations. 
From early statistical linear reductions (e.g., PCA, t-SNE~\cite{maaten2008visualizing}) to deep autoencoder-based non-linear embeddings (e.g., scDeepCluster~\cite{tian2019clustering}, scNAME~\cite{wan2022scname}), and further to structure-constrained learning (e.g., scGNN~\cite{wang2021scgnn} and scCDCG~\cite{xu2024sccdcg}), existing methods have achieved maturity in mining statistical distributions and topological structures.
However, these methods suffer from a fundamental limitation: \textit{semantic agnosticism}~\cite{cui2024scgpt,theodoris2023transfer}.
By reducing genes to abstract numerical indices, existing frameworks prioritize expression patterns over functional logic. Consequently, while adept at identifying distributional shifts, they lack the semantic grounding required for biological reasoning and interpretability \cite{xu2025scunified}.

To address this limitation, the field is witnessing a paradigm shift toward Foundation Models (FMs)~\cite{zhang2025survey}. 
Inspired by LLMs in NLP, these approaches analogize genes as "tokens" and cells as "sentences," pre-training on massive atlases to learn universal representations.
Pioneering works like scBERT~\cite{yang2022scbert} and Geneformer~\cite{theodoris2023transfer} adopted the BERT paradigm, capturing genome-wide context via masked modeling.
Subsequently, generative models such as scGPT~\cite{cui2024scgpt} and scFoundation~\cite{hao2024large} scaled up parameters to reconstruct complex expression patterns across species.
More recently, researchers have sought to enhance interpretability by integrating biological priors, such as gene regulatory networks in GeneCompass~\cite{yang2024genecompass} or leveraging textual embeddings in GenePT~\cite{chen2024genept}.

Despite their potential, directly adapting these general-purpose models to the specific downstream task of cell clustering presents structural contradictions~\cite{kedzierska2025zero,li2025sceval}:
(1) \textbf{Objective Mismatch between Pre-training and Clustering:}  Most models (e.g., scGPT) rely on "generative" pre-training focused on reconstructing global statistics rather than distinguishing local subpopulation boundaries. When applied to clustering (a discriminative task), this lack of boundary constraints often leads to "over-smoothing" representations, failing to separate subtle transitional states.
(2) \textbf{Numerical representations lack biological semantic support:} While treating genes as tokens, existing methods essentially capture numerical co-occurrence rather than explicit functional attributes (e.g., transcription factor regulation). Prioritizing probabilistic associations over causal logic makes them prone to hallucinations, thereby generating biologically implausible patterns driven by noise.
(3) \textbf{Missing Cross-Modal Alignment:} There is a lack of effective mechanisms to align "external biological knowledge" with "internal transcriptomic features" in the latent space. This modal disconnection results in clustering outputs that remain mathematical labels rather than biologically attributable entities~\cite{dip2025llm4cell}.

To address these impediments, we propose~\textbf{\methodname}, an \textbf{LLM}-Knowledge Enhanced Cross-Modal \textbf{D}eep \textbf{S}tructural \textbf{C}lustering framework for \textbf{s}ingle-\textbf{c}ell RNA Sequencing.
Diverging from the purely generative paradigm of general foundation models,~\methodname~is formulated as a task-specific \textit{knowledge-integrated} framework anchored in three core pillars: \textbf{Knowledge-Driven Priors, Structure-Aware Topology, and Cross-Modal Semantic Alignment}.
The core logic is to utilize LLMs not as data generators but as semantic mappers, explicitly integrating biological semantics from knowledge bases (e.g., NCBI) into the clustering process and enforcing alignment with numerical features via cross-modal contrastive learning.
\methodname~comprises three coupled modules:
(1) \textbf{Dual-Path Cell Semantic Encoding (Knowledge-driven):} It incorporates external priors by encoding NCBI gene summaries and using a Cell2Sentence strategy to capture dynamic regulation contexts.
(2) \textbf{Structure-Aware Cell Feature Encoding(Structure-aware):} Leveraging the scCDCG backbone, it extracts high-order topological information to ensure the fidelity of the native data manifold.
(3) \textbf{Cross-Modal Alignment and Fusion (Semantic-alignment):} Utilizing bidirectional InfoNCE loss and variance regularization, this module bridges the cognitive gap between "data-driven" structural signals and "knowledge-driven" semantic signals.
Extensive benchmarks demonstrate that~\methodname~significantly outperforms 11 mainstream baselines in both accuracy and efficiency, while uniquely endowing clustering results with explicit biological attribution.

Our contributions are summarized as follows:
\begin{itemize}
    \item \textbf{A Knowledge-Data Dual-Driven Paradigm:} 
    We transcend the limitations of traditional numerical indexing by systematically incorporating NCBI functional priors. This establishes a novel paradigm that bridges the gap between `structure-aware' deep learning and semantic-aware' biological reasoning.
    \item \textbf{Task-Specific Synergistic Optimization:} 
    To overcome the objective mismatch of FMs for clustering, we devise a specialized optimization objective. By integrating semantic alignment with discriminative structural constraints, we ensure the model focuses on defining precise cell subpopulation boundaries rather than global reconstruction.
    \item \textbf{Unified Interpretable Representation:} 
    We leverage cross-modal alignment to construct a unified latent space that preserves both topological fidelity and biological meaning, enabling interpretable mechanism discovery.
\end{itemize}

\section{Related Work}
\noindent\textbf{Conventional scRNA-seq Clustering.}
Early pipelines like Seurat~\cite{butler2018integrating} and Scanpy~\cite{wolf2018scanpy} rely on linear reduction and community detection. To capture non-linear manifolds and data sparsity, deep generative models such as scVI~\cite{lopez2018deep} and DCA~\cite{eraslan2019single} utilize ZINB-integrated autoencoders for explicit count modeling. Subsequent frameworks, including scDeepCluster~\cite{tian2019clustering} and scDCC~\cite{tian2021model}, introduced joint optimization of reconstruction and clustering-constrained objectives. 
To incorporate intercellular topological information, initial frameworks like scGNN~\cite{wang2021scgnn} and scDSC~\cite{gan2022deep} utilize graph convolutional networks for neighborhood aggregation, while scSiameseClu~\cite{xu2025scsiameseclu} employs contrastive learning to bolster representation robustness. 
To enhance scalability and circumvent GCN-induced over-smoothing, structural models based on deep graph-cut objectives have emerged. scCDCG~\cite{xu2024sccdcg} and its extension scSGC~\cite{xu2025soft} utilize deep spectral clustering to efficiently capture global topological constraints. Despite these advancements, these numerical-driven methods often lack explicit biological semantics, limiting their interpretability in complex tissues.


\noindent\textbf{Biological Foundation Models.}
The focus has recently shifted toward Transformer-based foundation models that learn universal biological representations from large-scale cell atlases. scBERT~\cite{yang2022scbert} and Geneformer~\cite{theodoris2023transfer} pioneered this paradigm by employing masked gene modeling to capture complex genomic contexts. To enhance generalization, scGPT~\cite{cui2024scgpt} and scFoundation~\cite{hao2024large} further scaled parameter capacities for multi-species generative embeddings.
Beyond sequence-only modeling, knowledge-enhanced frameworks such as GeneCompass~\cite{yang2024genecompass} and GenePT~\cite{chen2024genept} integrate gene regulatory networks and textual semantics, respectively, to infuse prior biological constraints. Recently, scMamba~\cite{yuan2025scmamba} was introduced to achieve linear computational complexity for ultra-long genomic sequences. Despite their powerful representation, these foundation models are not inherently optimized for partitioning tasks, suffering from a objective mismatch between general pre-training and downstream structural clustering.


\section{Methodology}









\begin{figure}[!t]
    \centering
    \includegraphics[width=1\linewidth]{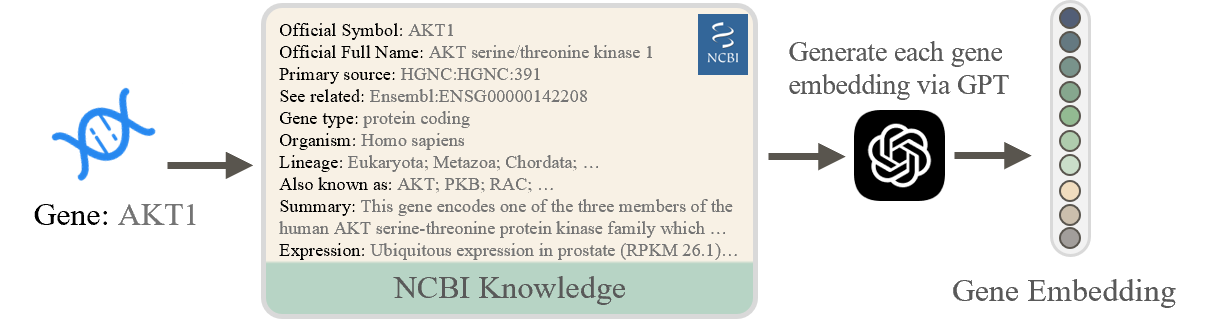}
    \caption{Details of Generating Gene Embedding Generation.}
    \label{fig:framework_GTG}
\end{figure}

\begin{figure*}[!th]
    \centering
    \includegraphics[width=0.8\textwidth]{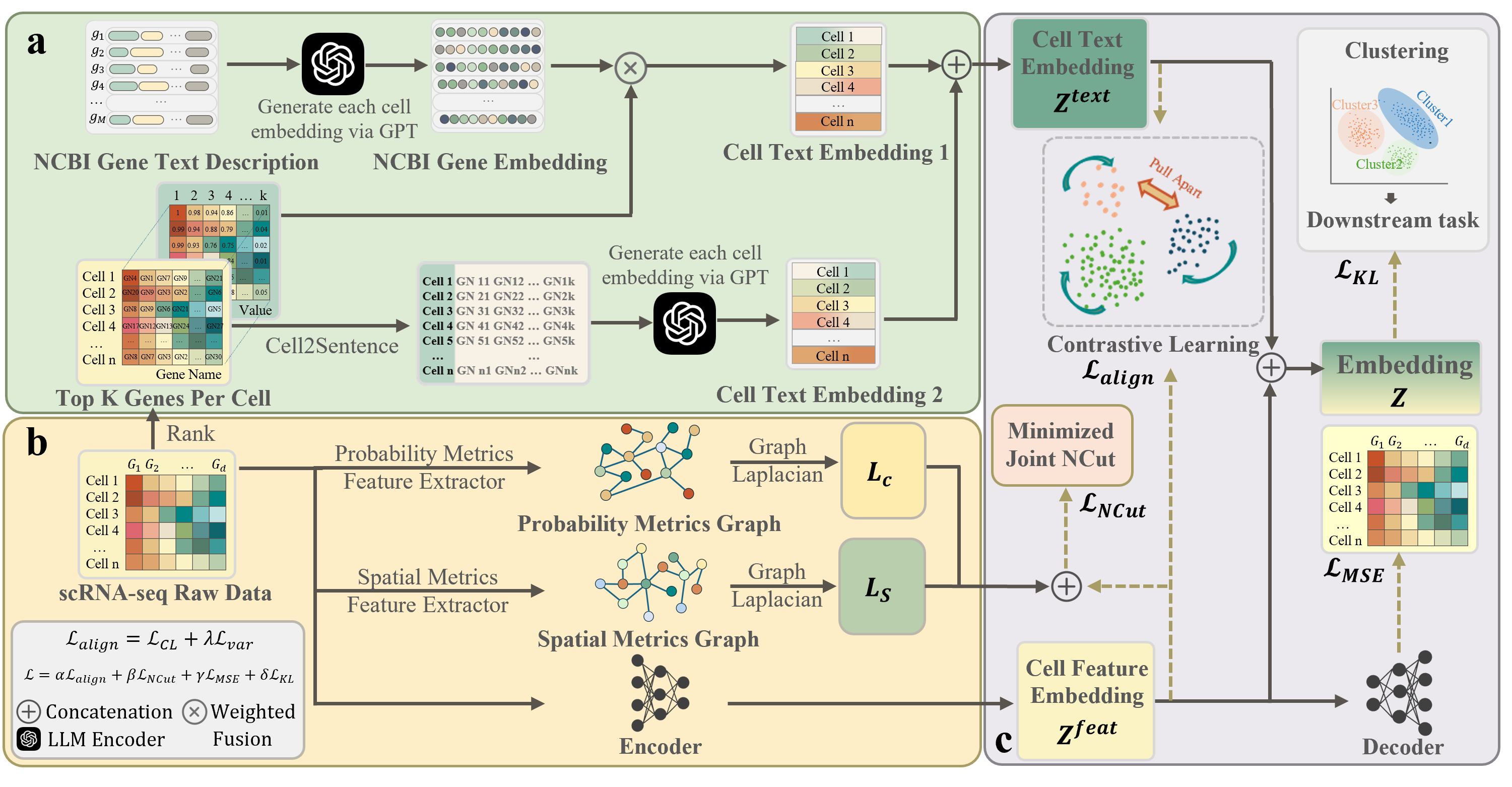}
    \caption{Framework Overview of the Proposed~\methodname. (a) Dual-path cell semantic encoding (Knowledge-driven); (b) Structure-aware cell feature encoding(Structure-aware); (c) Cross-modal alignment and fusion (Semantic-alignment).}
    \label{fig:framework}
\end{figure*}

\subsection{Problem Formulation}
The core challenge addressed in this study is \textbf{Knowledge-Enhanced Structural Clustering}. 
Let $\mathbf{X} \in \mathbb{R}^{N \times D}$ be the single-cell gene expression matrix, where $N$ denotes the number of cells and $D$ is the number of genes. 
Unlike traditional data-driven methods, we incorporate an external gene semantic knowledge base, denoted as $\mathbf{G}$, derived from LLMs.
Our objective is to learn a mapping function $f: (\mathbf{X}, \mathbf{G}) \to \mathbf{Z}$ that projects cells into a unified latent space $\mathbf{Z}$ by aligning transcriptomic structural features with biological semantic information. 
Ultimately, we aim to partition $\mathbf{Z}$ into $O$ cell populations, yielding the clustering assignment $\mathbf{C} = \{c_1, c_2, \dots, c_N\}$, where $c_i \in \{1, \dots, O\}$ denotes the cluster label of the $i$-th cell.

\subsection{Framework Overview}

The~\methodname~framework, illustrated in Fig.~\ref{fig:framework}, establishes a novel paradigm for LLM-Knowledge enhanced cross-modal deep structural clustering by synergizing large-scale biological priors with data-driven representation learning.
To integrate biological semantic knowledge, we first retrieve structured biological annotations from NCBI and encode them into gene-level textual representations using a frozen large language model. 
Subsequently, we construct a unified cell embedding through three modules: (1) \textbf{Semantic View}: creating complementary cell-level textual embeddings by aggregating gene semantics and serializing cell profiles; (2) \textbf{Structural View}: capturing the intrinsic data manifold via a graph-cut–guided deep structural encoder; and (3) \textbf{Cross-Modal Contrastive Fusion}: performing alignment between LLM-encoded biological semantics (Text) and expression-derived structural features (Feature), yielding a biologically grounded latent space optimized for the downstream clustering task.

\subsection{Gene Semantic Encoding}
To explicitly incorporate biological priors, we leverage a pre-trained LLM as a universal knowledge encoder, as illustrated in Fig.~\ref{fig:framework_GTG}. 
Specifically, we systematically collected comprehensive metadata covering genome-wide genes from the NCBI database~\cite{li2024screader}.
For the $j$-th gene ($j \in \{1, \dots, M\}$), this heterogeneous information (e.g., symbol, functional summary) is serialized into a structured prompt $\mathcal{T}_j$. 
We then extract the dense vector representation via the LLM encoder:
\begin{equation}
    \mathbf{g}_j = f_{\mathrm{LLM}}(\mathcal{T}_j).
    \label{equ:gene_text_embed}
\end{equation}
Consequently, we construct a global gene semantic matrix
\begin{equation}
    \mathbf{G} = [\, \mathbf{g}_1, \mathbf{g}_2, \dots, \mathbf{g}_M \,]^\top \in \mathbb{R}^{M \times d_1},
    \label{equ:Gene_text_embed}
\end{equation}
which projects discrete gene entities into a continuous, machine-interpretable semantic manifold.

\subsection{Dual-Path Cell Semantic Encoding}
To bridge the gap between gene-level knowledge and cell-level identity, we design a dual-path encoding strategy. Crucially, both paths operate on the most informative genes to filter noise and ensure computational efficiency.
For each cell, we strictly select the top-$K$ genes ranked by expression intensity (experimentally set to $K=2048$).
Let $\widetilde{\mathbf{X}} \in \mathbb{R}^{N \times K}$ denote this filtered and sorted expression matrix, where each row contains the expression values of the top-$K$ genes for a cell.

\noindent\textbf{Abundance-Weighted Semantic Aggregation.}
This branch explicitly utilizes the expression magnitudes in $\widetilde{\mathbf{X}}$ as importance weights.
Let $\widetilde{\mathbf{G}} \in \mathbb{R}^{K \times d_1}$ be the subset of LLM-derived gene embeddings corresponding to the genes in $\widetilde{\mathbf{X}}$.
The first cell-level embedding $\mathbf{Z}^{(1)}$ is obtained by aggregating these gene semantics weighted by their expression values:
\begin{equation}
    \mathbf{Z}^{(1)} = \widetilde{\mathbf{X}}\,\widetilde{\mathbf{G}} \in \mathbb{R}^{N \times d_1},
    \label{equ:cell_embed_1}
\end{equation}
which projects the cell's gene expression abundance directly into the semantic manifold defined by $\mathbf{G}$.

\noindent\textbf{Sequence-Based Contextual Modeling.}
To capture non-linear dependencies and gene co-occurrence patterns, we adopt a serialization approach inspired by Cell2Sentence~\cite{levine2024cell2sentence}.
We adopt a serialization approach where the row vector $\mathbf{\tilde{x}}_i$ from $\widetilde{\mathbf{X}}$ is treated as a sentence.
Specifically, the names of the top-$K$ genes (ordered by their rank in $\widetilde{\mathbf{X}}$) are tokenized into a sequence $\mathcal{S}_i$.
These sequences are then encoded by the shared LLM encoder $f_{\mathrm{LLM}}$:
\begin{equation}
    \mathbf{Z}^{(2)}_i = f_{\mathrm{LLM}}(\mathcal{S}_i), \quad \mathbf{Z}^{(2)} \in \mathbb{R}^{N \times d_1}.
    \label{equ:cell_embed_2}
\end{equation}
Unlike $\mathbf{Z}^{(1)}$, this representation leverages the LLM's attention mechanism to perceive the underlying regulatory logic of gene expressions preserved in the sorted gene list.

\noindent\textbf{Dual-Path Semantic Fusion.}
We integrate the abundance-based and context-based representations to yield the final unified cell semantic embedding $\mathbf{Z}^{\mathrm{text}}$:
\begin{equation}
    \mathbf{Z}^{\mathrm{text}} = \omega \mathbf{Z}^{(1)} + (1-\omega) \mathbf{Z}^{(2)},
    \label{equ:cell_text_final}
\end{equation}
where $\omega \in [0,1]$ is a balancing coefficient. This fused representation $\mathbf{Z}^{\mathrm{text}} \in \mathbb{R}^{N \times d_1}$ serves as the semantic anchor for the subsequent structure-aware alignment.

\subsection{Structure-Aware Cell Feature Encoding}

To capture the intrinsic topological manifold of the cell population, we utilize the deep graph clustering framework scCDCG~\cite{xu2024sccdcg} as our structural backbone.
Given input $\mathbf{X}$, the backbone functions as a non-linear mapping encoder $f_{\mathrm{struc}}(\cdot)$ to extract the cell feature embedding:
\begin{equation}
    \mathbf{Z}^{\mathrm{feat}} = f_{\mathrm{struc}}(\mathbf{X}) \in \mathbb{R}^{N \times d_2}.
    \label{equ:cell_feat_embed}
\end{equation}
This encoding process is governed by three complementary unsupervised objectives:
\begin{itemize}
    \item \textbf{Topology Preservation ($\mathcal{L}_{NCut}$):} A differentiable Normalized Cut loss that captures high-order structural information via long-range dependencies, effectively preserving the global manifold while circumventing the over-smoothing limitations of traditional GCNs.
    \item \textbf{Expression Fidelity ($\mathcal{L}_{MSE}$):} An autoencoder-based reconstruction loss that guarantees the latent embeddings retain essential gene expression information.
    \item \textbf{Cluster Refinement ($\mathcal{L}_{KL}$):} A KL-divergence objective that aligns soft assignments with a target distribution optimized via Optimal Transport (OT). This mechanism enforces a global balance constraint to prevent cluster collapsing and stabilize the partitioning manifold.

\end{itemize}

The resulting $\mathbf{Z}^{\mathrm{feat}}$ encapsulates the data-driven structural priors, serving as the topological anchor for the subsequent cross-modal alignment.

\subsection{Cross-Modal Alignment and Fusion}
To bridge the heterogeneity between biological semantics and topological features, we align the semantic embedding $\mathbf{Z}^{\mathrm{text}}$ and the structural embedding $\mathbf{Z}^{\mathrm{feat}}$ into a unified latent space via contrastive learning.

\noindent\textbf{Projection and Similarity.}
We first map the modality-specific embeddings to a shared $d$-dimensional space using two separate non-linear projection heads (two-layer MLPs with ReLU), denoted as $f_{\phi}(\cdot)$ and $g_{\psi}(\cdot)$:
\begin{equation}
    \mathbf{\hat{Z}}^{\mathrm{text}} = f_{\phi}(\mathbf{Z}^{\mathrm{text}}), \quad 
    \mathbf{\hat{Z}}^{\mathrm{feat}} = g_{\psi}(\mathbf{Z}^{\mathrm{feat}}),
\end{equation}
where $\mathbf{\hat{Z}} \in \mathbb{R}^{N \times d}$. 
We then compute the cross-modal cosine similarity matrix $\mathbf{S} \in \mathbb{R}^{N \times N}$. 
By normalizing the projected embeddings to the unit hypersphere (i.e., $\|\mathbf{\hat{z}}\| = 1$), the similarity is efficiently calculated via scaled dot product:

\begin{equation} 
    \mathbf{S} = (\mathbf{\hat{Z}}^{\mathrm{text}} (\mathbf{\hat{Z}}^{\mathrm{feat}})^\top) / \tau,
\end{equation}
where $\tau$ is a temperature parameter.

\noindent\textbf{Bidirectional Alignment Objective.}
To enforce semantic consistency, we minimize the symmetric InfoNCE loss. 
For a mini-batch of $N$ cells, considering the paired representations $(i, i)$ as positives and all other pairs as negatives, the loss is formulated as:
\begin{equation}
\small
    \mathcal{L}_{CL} = -\frac{1}{2N} \sum_{i=1}^{N} \left( 
    \underbrace{\log \frac{\exp(S_{ii})}{\sum_{k=1}^{N} \exp(S_{ik})}}_{\text{Text } \to \text{ Feature}} + 
    \underbrace{\log \frac{\exp(S_{ii})}{\sum_{k=1}^{N} \exp(S_{ki})}}_{\text{Feature } \to \text{ Text}} 
    \right).
\end{equation}
The first term aligns the textual view to the structural view, while the second aligns the structural view to the textual view.

\noindent\textbf{Variance Regularization.}
To prevent the projection heads from mapping diverse cells to a trivial point (i.e., dimensional collapse), we impose a hinge loss on the standard deviation of each feature dimension:
\begin{equation}
\small
    \mathcal{L}_{var} = \sum_{\mathbf{V} \in \{\mathbf{\hat{Z}}^{\mathrm{text}}, \mathbf{\hat{Z}}^{\mathrm{feat}}\}} \frac{1}{d} \sum_{k=1}^{d} \max\left(0, 1 - \sqrt{\mathrm{Var}(\mathbf{V}_{:,k}) + \epsilon}\right),
\end{equation}
where $\mathbf{V}_{:,k} \in \mathbb{R}^{N}$ denotes the column vector corresponding to the $k$-th dimension across the batch, and $\epsilon=10^{-4}$ is a constant for numerical stability.


The overall loss for the cross-modal alignment module is formulated as:
\begin{equation}
    \mathcal{L}_{align} = \mathcal{L}_{CL} + \lambda \mathcal{L}_{var},
\end{equation}
where $\lambda$ is a hyperparameter.

\subsection{Optimization and Inference}
 
\noindent\textbf{Joint Training.}
The framework is optimized end-to-end via a weighted composite objective:
\begin{equation}
    \mathcal{L} = \alpha \mathcal{L}_{align} + \beta \mathcal{L}_{NCut} + \gamma \mathcal{L}_{MSE} + \delta \mathcal{L}_{KL},
    \label{equ:loss_all}
\end{equation}
where $\alpha, \beta, \gamma$, and $\delta$ are hyperparameters balancing cross-modal semantic alignment, global topology preservation, reconstruction fidelity, and cluster refinement, respectively.

\noindent\textbf{Inference.}
Upon convergence, we obtain the unified embeddings by averaging the aligned semantic and structural views:
\begin{equation}
    \mathbf{Z}^{\mathrm{cluster}} = \frac{1}{2} \left( \mathbf{\hat{Z}}^{\mathrm{text}} + \mathbf{\hat{Z}}^{\mathrm{feat}} \right).
\end{equation}
The final clustering assignments $\mathbf{C}$ are then generated by applying K-Means (or Leiden) directly to $\mathbf{Z}^{\mathrm{cluster}}$.

\begin{table}[!t]
    \centering
    \scriptsize
    \setlength{\tabcolsep}{4pt}
    \renewcommand{\arraystretch}{1}
    \begin{tabular}{llcccc}
    \toprule
    \textbf{Species} & \textbf{Dataset Name} & \textbf{\#Cell} & \textbf{\#Gene} & \textbf{\#Cluster} & \textbf{Sparsity (\%)} \\
    \midrule
    \multirow{3}{*}{\textbf{\makecell{Human}}} 
    & Mauro Pancreas & 2,122 & 19,046 & 9 & 73.02 \\
    & Sonya Liver & 8,444 & 4,999 & 11 & 90.77 \\
    & Sapiens Liver & 2,152 & 61,759 & 15 & 95.42 \\
    \midrule
    \multirow{3}{*}{\textbf{\makecell{Mouse}  }} 
    & Muris Brain & 13,417 & 21,609 & 2 & 91.83 \\
    & Muris Liver & 2,859 & 21,609 & 11 & 88.20 \\
    & Muris Limb Muscle & 3,855 & 21,609 & 6 & 91.38 \\
    \bottomrule
    \end{tabular}
    \caption{Details of single-cell RNA-seq Datasets.} 
    \label{tab:datasets}
\end{table}

\begin{table*}[!th]
\resizebox{\textwidth}{!}{%
\centering
\footnotesize 
\renewcommand{\arraystretch}{0.88}  
\setlength{\extrarowheight}{0pt}      
\begin{tabular}{l|c|cccccccccc}
\toprule
\multirow{3}{*}{\textbf{Dataset}} & 
\multirow{3}{*}{\textbf{Metrics}} & 
\multicolumn{4}{c}{\textbf{Deep Non-linear Embedding Models}} & 
& \multicolumn{5}{c}{\textbf{Deep Structural Clustering Models}} \\
\cmidrule{3-6} \cmidrule{8-12}
& & scDeepCluster & scMAE & scNAME & scziDesk & 
& scGNN & scDSC & scSiameseClu & scCDCG & \textbf{OURS}\\
\midrule

\multirow{3}*{\makecell{Mauro\\Pancreas}}
 & ACC
 & $73.55_{\pm1.37}$ & $95.62_{\pm0.16}$ & $89.73_{\pm9.92}$ & $69.97_{\pm17.08}$ & 
 & $79.32_{\pm3.96}$ & $79.42_{\pm1.34}$ & $\underline{94.95}_{\pm0.15}$ & $92.65_{\pm2.93}$ & $\textbf{96.94}_{\pm1.00}$ \\
 & NMI
 & $78.89_{\pm0.98}$ & $88.49_{\pm0.44}$ & $85.31_{\pm5.11}$ & $64.94_{\pm14.21}$ & 
 & $78.76_{\pm5.60}$ & $75.89_{\pm0.61}$ & $86.19_{\pm0.34}$ & $\underline{86.81}_{\pm0.98}$ & $\textbf{92.43}_{\pm1.49}$ \\
 & ARI
 & $64.90_{\pm1.07}$ & $92.38_{\pm0.35}$ & $84.51_{\pm13.42}$ & $50.31_{\pm29.92}$ & 
 & $78.81_{\pm3.17}$ & $75.42_{\pm1.22}$ & $\underline{94.59}_{\pm0.35}$ & $91.37_{\pm1.21}$ & $\textbf{96.08}_{\pm0.67}$ \\
 \hline
 
\multirow{3}*{\makecell{Sonya\\liver}}
 & ACC
 & $69.21_{\pm2.47}$ & $80.73_{\pm1.86}$ & $79.97_{\pm8.80}$ & $76.58_{\pm1.52}$ & 
 & $72.33_{\pm3.40}$ & $70.90_{\pm2.23}$ & $\underline{88.33}_{\pm1.74}$ & $75.34_{\pm3.67}$ & $\textbf{90.47}_{\pm6.44}$ \\
 & NMI
 & $79.69_{\pm0.68}$ & $85.59_{\pm1.44}$ & $85.04_{\pm3.65}$ & $83.34_{\pm0.70}$ & 
 & $73.56_{\pm2.70}$ & $71.63_{\pm2.75}$ & $\underline{88.82}_{\pm1.24}$ & $79.34_{\pm2.59}$ & $\textbf{91.92}_{\pm9.22}$ \\
 & ARI
 & $69.71_{\pm1.18}$ & $88.92_{\pm1.78}$ & $83.96_{\pm13.63}$ & $83.87_{\pm1.17}$ & 
 & $76.01_{\pm2.60}$ & $75.34_{\pm3.56}$ & $\underline{91.90}_{\pm0.48}$ & $81.26_{\pm2.69}$ & $\textbf{92.01}_{\pm5.58}$ \\
 \hline
 
\multirow{3}*{\makecell{Sapiens\\Liver}}
 & ACC
 & $49.08_{\pm1.90}$ & $67.51_{\pm2.30}$ & $63.33_{\pm1.70}$ & $68.19_{\pm0.60}$ & 
 & $\underline{73.83}_{\pm3.00}$ & $64.67_{\pm4.40}$ & $62.34_{\pm0.84}$ & $73.09_{\pm1.40}$ & $\textbf{75.19}_{\pm2.82}$ \\
 & NMI
 & $60.98_{\pm2.20}$ & $\underline{78.25}_{\pm0.60}$ & $74.83_{\pm0.70}$ & $75.10_{\pm1.20}$ & 
 & $77.21_{\pm2.00}$ & $68.21_{\pm5.20}$ & $68.96_{\pm1.60}$ & $62.54_{\pm3.20}$ & $\textbf{79.09}_{\pm5.97}$ \\
 & ARI
 & $34.52_{\pm2.00}$ & $63.82_{\pm4.10}$ & $58.94_{\pm2.50}$ & $66.12_{\pm1.90}$ & 
 & $\underline{72.36}_{\pm1.70}$ & $56.94_{\pm7.90}$ & $51.10_{\pm1.29}$ & $41.11_{\pm4.40}$ & $\textbf{78.03}_{\pm0.97}$ \\

 \hline
 
\multirow{3}*{\makecell{Muris\\Limb\\Muscle}}
 & ACC
 & $59.57_{\pm3.90}$ & $66.13_{\pm3.40}$ & $61.34_{\pm3.10}$ & $53.31_{\pm4.30}$ & 
 & $48.62_{\pm2.30}$ & $64.37_{\pm4.00}$ & $81.32_{\pm6.88}$ & $\underline{94.50}_{\pm7.10}$ & $\textbf{95.54}_{\pm3.86}$\\
 & NMI
 & $34.55_{\pm4.90}$ & $59.44_{\pm3.80}$ & $63.51_{\pm1.30}$ & $36.66_{\pm6.40}$ & 
 & $22.10_{\pm4.70}$ & $33.30_{\pm4.50}$ & $\underline{80.04}_{\pm4.44}$ & $56.54_{\pm7.60}$ & $\textbf{91.69}_{\pm8.60}$ \\
 & ARI
 & $35.95_{\pm8.10}$ & $51.54_{\pm3.30}$ & $54.70_{\pm2.30}$ & $32.59_{\pm7.20}$ & 
 & $21.90_{\pm3.20}$ & $34.28_{\pm6.40}$ & $\underline{75.49}_{\pm8.34}$ & $53.37_{\pm8.50}$ & $\textbf{82.91}_{\pm5.60}$ \\
 \hline
 
\multirow{3}*{\makecell{Muris\\Brain}}
 & ACC
 & $85.36_{\pm18.10}$ & $71.37_{\pm0.00}$ & $90.24_{\pm0.30}$ & \textcolor{gray}{OOM} & 
 & $91.40_{\pm0.10}$ & $96.02_{\pm2.50}$ & $\underline{99.17}_{\pm5.19}$ & $95.55_{\pm1.10}$ & $\textbf{99.42}_{\pm0.51}$ \\
 & NMI
 & $\underline{59.48}_{\pm44.40}$ & $1.33_{\pm0.00}$ & $0.01_{\pm0.00}$ & \textcolor{gray}{OOM} & 
 & $0.31_{\pm0.10}$ & $0.07_{\pm0.00}$ & $33.01_{\pm1.79}$ & $22.48_{\pm8.30}$ & $\textbf{68.58}_{\pm1.25}$ \\
 & ARI
 & $\underline{60.11}_{\pm46.80}$ & $-2.22_{\pm0.00}$ & $0.34_{\pm0.10}$ & \textcolor{gray}{OOM} & 
 & $2.77_{\pm0.50}$ & $-0.37_{\pm0.80}$ & $30.87_{\pm1.22}$ & $35.56_{\pm7.80}$ & $\textbf{79.36}_{\pm1.96}$ \\
 \hline
 
\multirow{3}*{\makecell{Muris\\Liver}}
 & ACC
 & $42.62_{\pm3.20}$ & $53.48_{\pm0.40}$ & $49.72_{\pm4.10}$ & $44.50_{\pm3.40}$ & 
 & $51.58_{\pm2.90}$ & $55.76_{\pm7.50}$ & $66.81_{\pm4.50}$ & $\underline{68.13}_{\pm1.40}$ & $\textbf{75.24}_{\pm6.58}$ \\
 & NMI
 & $45.07_{\pm2.20}$ & $65.39_{\pm1.00}$ & $61.21_{\pm1.70}$ & $43.17_{\pm3.40}$ & 
 & $55.15_{\pm1.10}$ & $35.07_{\pm4.70}$ & $\underline{71.46}_{\pm1.51}$ & $62.06_{\pm2.40}$ & $\textbf{88.38}_{\pm6.54}$ \\
 & ARI
 & $27.92_{\pm3.90}$ & $47.55_{\pm0.60}$ & $38.55_{\pm3.40}$ & $27.27_{\pm4.60}$ & 
 & $45.58_{\pm3.20}$ & $43.74_{\pm12.30}$ & $\underline{49.56}_{\pm3.16}$ & $46.96_{\pm3.70}$ & $\textbf{69.85}_{\pm1.99}$ \\

 \hline

\multirow{3}*{AVG}
 & ACC
 & $63.23_{\pm5.16}$ & $72.47_{\pm1.35}$ & $72.39_{\pm4.65}$ & $62.51_{\pm5.38}$ & 
 & $69.51_{\pm2.61}$ & $71.85_{\pm3.66}$ & $82.15_{\pm3.22}$ & $\underline{83.21}_{\pm2.93}$ & $\textbf{88.80}_{\pm3.54}$ \\
 & NMI
 & $59.78_{\pm9.23}$ & $63.08_{\pm1.21}$ & $61.65_{\pm2.08}$ & $60.64_{\pm5.18}$ & 
 & $51.18_{\pm2.70}$ & $46.03_{\pm2.96}$ & $\underline{71.41}_{\pm1.82}$ & $61.63_{\pm4.18}$ & $\textbf{85.35}_{\pm5.51}$ \\
 & ARI
 & $48.85_{\pm10.50}$ & $57.00_{\pm1.69}$ & $53.50_{\pm5.89}$ & $50.03_{\pm7.45}$ & 
 & $49.44_{\pm2.40}$ & $47.55_{\pm5.36}$ & $\underline{65.59}_{\pm2.47}$ & $58.27_{\pm4.72}$ & $\textbf{83.04}_{\pm2.80}$ \\
\hline

RANK & ALL & 6 & 4 & 5 & 6 & & 9 & 8 & 2 & 3 & 1 \\
\bottomrule
\end{tabular}}
\caption{Clustering performance compared with deep embedding and structural clustering models (mean$_{\pm \text{std}}$). Bold and \underline{underline} denote best/runner-up. (\textcolor{gray}{OOM}: Out-of-Memory; Rank: avg. of ACC/NMI/ARI; scDeepCluster and scziDesk tied for 6th).}
\label{tab:clustering_performance_dl}
\end{table*}

\begin{figure*}[!th]
\centering
\subfloat[scDeepCluster]{
\includegraphics[width=0.14\textwidth]{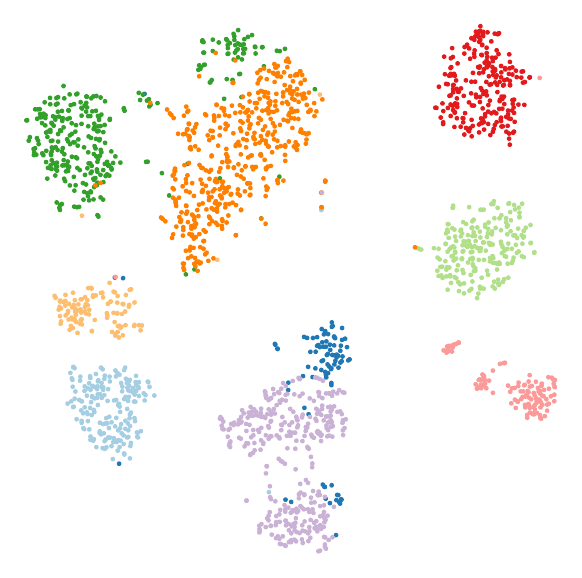}
}
\subfloat[scMAE]{
\includegraphics[width=0.14\textwidth]{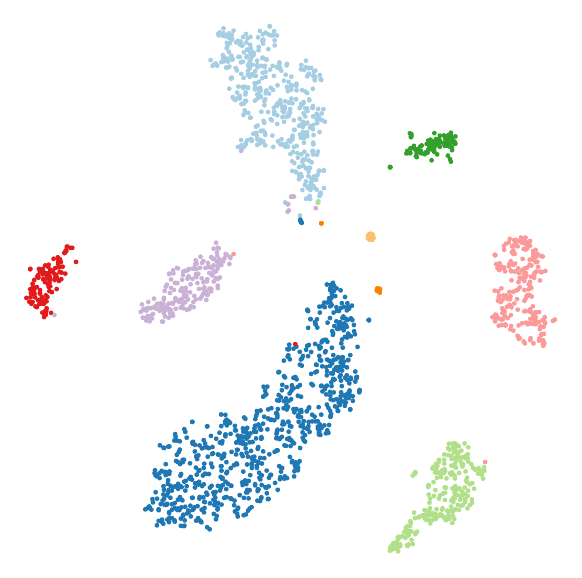}
}
\subfloat[scSiameseClu]{
\includegraphics[width=0.14\textwidth]{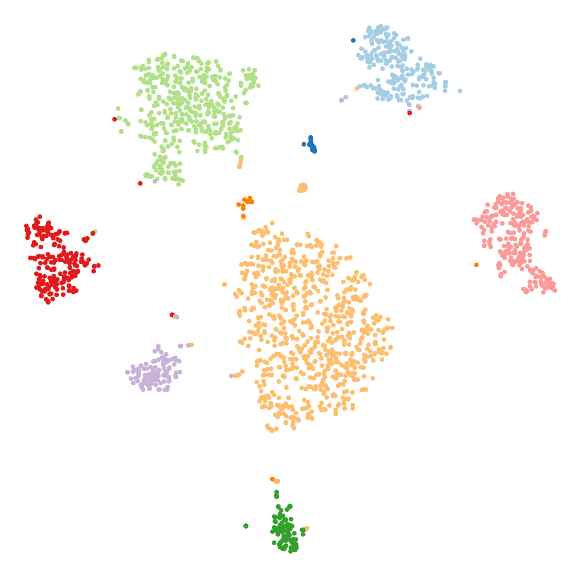}
}
\subfloat[scCDCG]{
\includegraphics[width=0.14\textwidth]{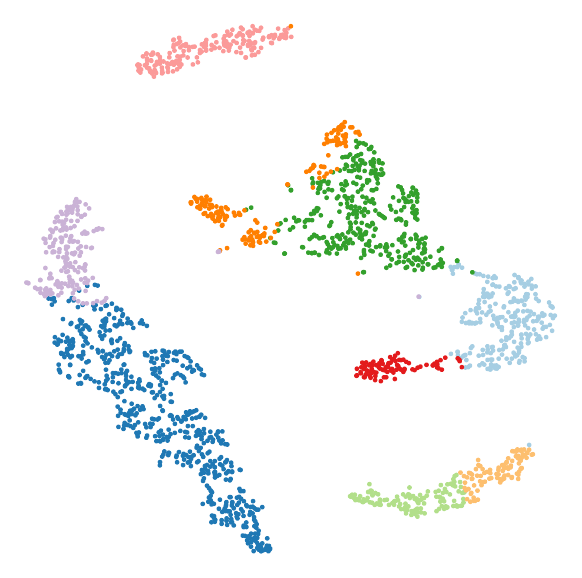}
}
\subfloat[\methodname]{
\includegraphics[width=0.14\textwidth]{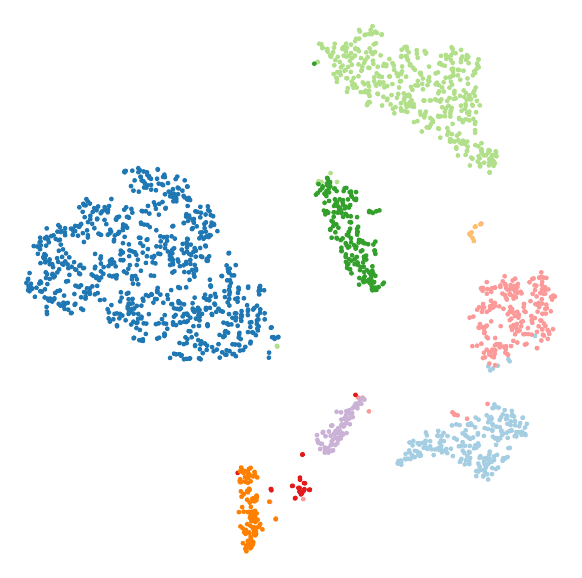}
}
\caption{Visualization of~\methodname~and four typical baselines on \textit{Mauro Pancreas} in 2D t-SNE projection. }
\label{fig:visualization}
\end{figure*}

\section{Experiments}  
\subsection{Experimental Setup}

\noindent\textbf{Dataset.}  
We utilized a subset of datasets provided by scCluBench~\cite{xu2025scclubench}, consisting of 6 real scRNA-seq datasets from human and mouse tissues (Table~\ref{tab:datasets}). All data were used as provided, without further preprocessing.


\noindent\textbf{Baseline Methods.} 
We compare~\methodname~against 11 baselines across three categories:
(1) Deep Non-linear Embedding Models: These models utilize deep autoencoders to learn latent representations, including scDeepCluster~\cite{tian2019clustering}, scziDesk~\cite{chen2020deep}, and masked autoencoder variants scNAME~\cite{wan2022scname} and scMAE~\cite{fang2024scmae}. 
(2) Deep Structural Clustering Models: These methods explicitly incorporate cellular topological information via graph-based architectures, comprising scGNN~\cite{wang2021scgnn}, scDSC~\cite{gan2022deep}, the contrastive scSiameseClu~\cite{xu2025scsiameseclu}, and the cut-informed scCDCG~\cite{xu2024sccdcg}. 
(3) Foundation Models: These large-scale pre-trained frameworks leverage expansive atlases for downstream transfer, represented by scGPT~\cite{cui2024scgpt}, GeneFormer~\cite{theodoris2023transfer}, and the knowledge-informed GeneCompass~\cite{yang2024genecompass}. 
All methods were implemented using their publicly available Python packages, with hyperparameters tuned for optimal performance.

\noindent\textbf{Implementation Details.}
Implemented in Python 3.12 (PyTorch $\ge$2.9.1, CUDA 13.0), ~\methodname~follows a two-stage protocol: autoencoder pre-training (joint NCut/MSE) followed by contrastive clustering. 
We generate gene embeddings from NCBI metadata using OpenAI's \textit{text-embedding-3-small}. 
Cell embeddings are extracted from NCBI metadata via OpenAI's \textit{text-embedding-3-small}. Cell embeddings aggregate the top-2048 highly expressed genes ($k=2048$) weighted by expression $v_i$: $\mathbf{e}_{\text{cell}} = \sum_{i=1}^{k} v_i \cdot \mathbf{e}_{\text{gene}_i}$. 
The architecture features a [256, 16] autoencoder producing 16-dimensional latent vectors. A 128-dimensional projection head aligns modalities via bi-directional InfoNCE loss ($\tau=0.1$) with variance regularization ($\lambda=10^{-2}$).
Training spans 200 epochs/stage using Adam, with hyperparameters optimized via Optuna. Clusters are initialized by KMeans and refined via Sinkhorn ($\lambda=5$, 1k iters). We report the mean of 5 independent runs on an NVIDIA A800-80GB GPU.
All datasets and code for~\methodname~are available at the link: 
\url{https://github.com/XPgogogo/scLLM-DSC}.


\noindent\textbf{Evaluation Metrics.} We assess the clustering quality by measuring the congruence between predicted partitions and biological ground truth using three benchmark indices: Clustering Accuracy (ACC), Normalized Mutual Information (NMI), and Adjusted Rand Index (ARI). 
While ACC and NMI range from 0 to 1, ARI can be negative if clustering performs worse than random assignment. 
For all metrics, higher values indicate superior clustering fidelity.

\subsection{Overall Performance}

\begin{table}[!t]
  \centering
  \resizebox{0.48\textwidth}{!}{%
  \footnotesize
  \renewcommand{\arraystretch}{0.85}  
  \setlength{\extrarowheight}{0pt}      
  \renewcommand{\arraystretch}{1}
  \begin{tabular}{lccccc}
    \toprule
    \textbf{Dataset} & \textbf{Metric} & \textbf{scGPT} & \textbf{Geneformer} & \textbf{GeneCompass} & \textbf{OURS} \\
    \midrule
    \multirow{3}{*}{\makecell[l]{Mauro\\Pancreas}} 
      & ACC & $56.61_{\pm1.35}$ & $17.14_{\pm0.67}$ & $25.19_{\pm0.95}$ & $\mathbf{96.94}_{\pm1.00}$ \\
      & NMI & $53.91_{\pm1.26}$ & $0.87_{\pm0.15}$ & $7.00_{\pm0.11}$ & $\mathbf{92.43}_{\pm1.49}$ \\
      & ARI & $37.15_{\pm1.49}$ & $0.02_{\pm0.04}$ & $3.00_{\pm0.19}$ & $\mathbf{96.08}_{\pm0.67}$ \\
    \hline
    
    \multirow{3}{*}{\makecell[l]{Sonya\\Liver}} 
      & ACC & $57.75_{\pm8.35}$ & $\mathbf{100.00}_{\pm0.00}$ & $\mathbf{100.00}_{\pm0.00}$ & $90.47_{\pm6.44}$ \\
      & NMI & $76.47_{\pm3.40}$ & $\mathbf{100.00}_{\pm0.00}$ & $\mathbf{100.00}_{\pm0.00}$ & $91.92_{\pm9.22}$ \\
      & ARI & $53.74_{\pm9.45}$ & $\mathbf{100.00}_{\pm0.00}$ & $\mathbf{100.00}_{\pm0.00}$ & $92.01_{\pm5.58}$ \\
    \hline
    
    \multirow{3}{*}{\makecell[l]{Sapiens\\Liver}} 
      & ACC & $43.47_{\pm3.91}$ & $24.49_{\pm0.98}$ & $27.77_{\pm1.31}$ & $\mathbf{75.19}_{\pm2.82}$ \\
      & NMI & $59.75_{\pm1.25}$ & $23.78_{\pm0.78}$ & $24.23_{\pm0.78}$ & $\mathbf{79.09}_{\pm5.97}$ \\
      & ARI & $36.11_{\pm6.46}$  & $10.75_{\pm0.64}$ & $13.48_{\pm2.19}$ & $\mathbf{78.03}_{\pm0.97}$ \\
    \hline

    \multirow{3}{*}{\makecell[l]{Muris\\Limb Muscle}} 
      & ACC & $29.22_{\pm0.25}$ & $23.25_{\pm1.15}$ & $24.47_{\pm0.04}$ & $\mathbf{95.54}_{\pm3.86}$ \\
      & NMI & $9.44_{\pm0.06}$ & $1.05_{\pm0.20}$ & $4.16_{\pm0.01}$ & $\mathbf{91.69}_{\pm8.60}$ \\
      & ARI & $5.64_{\pm0.31}$ & $0.77_{\pm0.30}$ & $2.40_{\pm0.00}$ & $\mathbf{82.91}_{\pm5.60}$ \\

    \hline
    \multirow{3}{*}{\makecell[l]{Muris\\Brain}} 
      & ACC & $59.54_{\pm0.57}$ & $62.71_{\pm0.10}$ & $54.82_{\pm0.02}$ & $\mathbf{99.42}_{\pm0.51}$ \\
      & NMI & $0.04_{\pm0.02}$ & $0.02_{\pm0.00}$ & $0.00_{\pm0.00}$ & $\mathbf{68.58}_{\pm1.25}$  \\
      & ARI & $0.20_{\pm0.05}$ & $0.17_{\pm0.00}$ & $-0.01_{\pm0.00}$ & $\mathbf{79.36}_{\pm1.96}$ \\

    \hline
    \multirow{3}{*}{\makecell[l]{Muris\\Liver}} 
      & ACC & $32.44_{\pm2.20}$ & $13.84_{\pm0.45}$ & $28.80_{\pm2.44}$ & $\mathbf{75.24}_{\pm6.58}$ \\
      & NMI & $29.40_{\pm1.25}$ & $2.01_{\pm0.15}$ & $19.70_{\pm0.79}$ & $\mathbf{83.88}_{\pm6.54}$ \\
      & ARI & $13.61_{\pm0.44}$ & $-0.03_{\pm0.17}$ & $13.48_{\pm2.16}$ & $\mathbf{69.85}_{\pm1.99}$ \\

    \bottomrule
  \end{tabular}}
  \caption{Clustering performance comparison with biological foundation models (mean$_{\pm \text{std}}$). The \textbf{bold} values represent the best results.}
  \label{tab:clustering_performance_fm}
\end{table}

\noindent\textbf{Qualitative Analysis.} 
 (1)\textit{\textbf{Comparison with Deep Clustering Baselines.}}
Tab.~\ref{tab:clustering_performance_dl} shows that \methodname~achieves \textbf{SOTA performance}, outperforming all baselines in both average metrics (ACC: 88.80\%, NMI: 85.35\%, ARI: 83.04\%) and overall ranking. 
The superiority stems from its ability to synergize LLM-derived semantic anchors with global structural manifolds. 
Specifically, the performance margin over \textit{deep non-linear embedding models} underscores the necessity of integrating both textual semantics and topological structures. Furthermore, compared to \textit{deep structural clustering models} (e.g., scCDCG), \methodname~yields higher clustering fidelity, validating that LLM-based priors provide more discriminative guidance for structural alignment than raw numerical features.
This robustness is further exemplified on the \textit{Muris Brain} dataset (\#Cell=13,417, \#Cluster=2). 
While many baselines (e.g., scziDesk) achieve misleadingly high ACC by predicting majority classes, their near-zero NMI and ARI indicate a collapse into trivial partitions.
Such failure occurs because raw numerical features often lack sufficient margin to define clear boundaries in high-volume but low-cluster scenarios, causing structural models to suffer from oscillation or over-smoothing.
~\methodname~overcomes this by injecting LLM-based semantic anchors, which provide the contrastive signals required to maintain cluster separation and prevent manifold collapse.
(2) \textit{\textbf{Comparison with Biological Foundation Models.}} 
Evaluation against large-scale foundation models (Tab.~\ref{tab:clustering_performance_fm}) reveals that \methodname~provides more robust partitioning across most datasets. 
Unlike general-purpose models, \methodname~is specialized for clustering tasks, effectively capturing discrete cell-type boundaries. 
While Geneformer and GeneCompass show localized peaks on \textit{Sonya Liver}, likely due to pre-training data overlap, \methodname~demonstrates superior generalization elsewhere. 
This strongly underscores that for effectively resolving highly complex cellular heterogeneity, dedicated structural objectives specifically tailored for clustering are indispensable compared to universal embeddings.

\begin{figure*}[!t]
    \centering
    \includegraphics[width=1\linewidth]{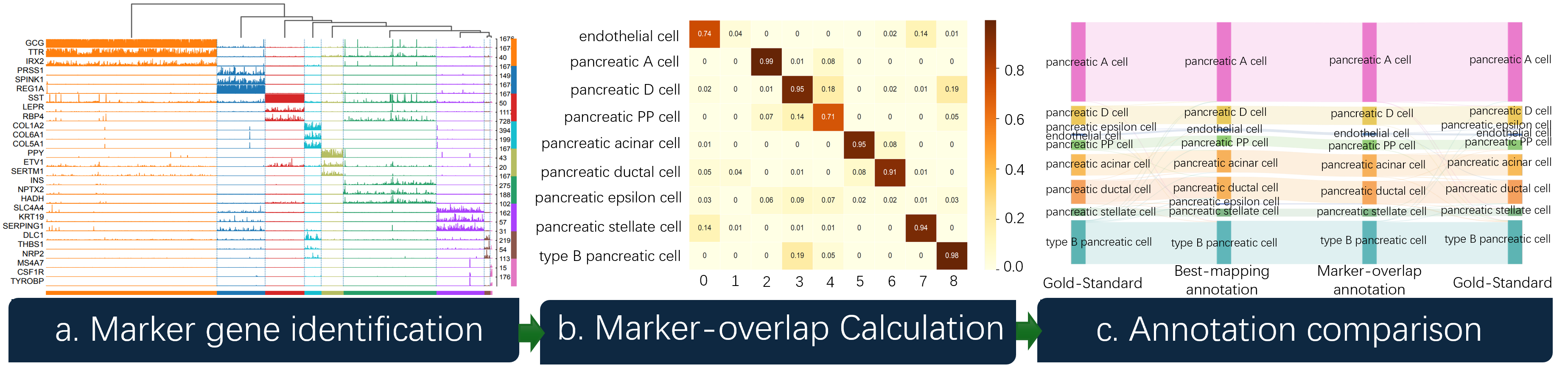}
    \caption{All Biological Analysis of~\methodname on \textit{Mauro Pancreas} Dataset.}
    \label{fig:Bio_scLLM-DSC_all}
\end{figure*}

\noindent\textbf{Quantitative Analysis.} 
As depicted in Fig.~\ref{fig:visualization}, we project the latent representations of~\methodname~and four representative baselines into a 2D space. The visualization results demonstrate that~\methodname~yields substantial advantages in both boundary clarity and intra-cluster compactness. Regarding the cluster boundaries, baselines such as scDeepCluster and scCDCG exhibit discernible "bridge" structures, resulting in marginal overlaps between distinct cell populations. Conversely,~\methodname~effectively eliminates such structural ambiguity; by leveraging LLM-based semantic priors to provide robust discriminative signals, it defines sharp and clean biological boundaries. Notably, from the perspective of intra-cluster compactness,~\methodname~facilitates a higher degree of aggregation within the manifold space, forming a more dense and robust cluster structure than the baselines. 
This dual enhancement in boundary separation and compactness validates the efficacy of cross-modal alignment in constraining the latent space to capture fine-grained cellular heterogeneity.




\subsection{Biological Analysis}
We conduct an in-depth analysis of the biological relevance of our clustering results following the standard evaluation pipeline of scCluBench~\cite{xu2025scclubench}. 
Taking the \textit{Mauro Pancreas} dataset as a case study, we first identify the top 100 marker genes for each predicted cluster via SCANPY, with their expression specificity visualized in the tracksplot in Fig.~\ref{fig:Bio_scLLM-DSC_all}(a). 
Subsequently, Fig.~\ref{fig:Bio_scLLM-DSC_all}(b) illustrates the marker-overlap calculation process, which establishes a statistical foundation for biological identity assignments by calculating the intersection (overlap score) between predicted markers and gold-standard reference sets.
Furthermore, in the Sankey diagram of Fig.~\ref{fig:Bio_scLLM-DSC_all}(c), we compare two annotation paradigms: the purely numerical \textit{Best-Mapping} (BM) strategy, which operates independently of external biological priors, and the biology-guided \textit{Marker-Overlap} (MO) strategy. 
Analytical results demonstrate that while the BM strategy tends to "enforce mappings" for all nine potential cell types, the MO strategy exhibits superior biological rigor. 
Concretely, for cluster 1 in Fig.~\ref{fig:Bio_scLLM-DSC_all}(b) mislabeled as "\textit{ epsilon cell}" by BM, the MO strategy deliberately labels the cluster as "\textit{pancreatic ductal cell}" according to its highest overlap score, as shown in Fig.~\ref{fig:Bio_scLLM-DSC_all}(c), ensuring higher fidelity cellular identification.





\subsection{Ablation Study}

\begin{figure}[t!]
    \centering
    \subfloat[Framework Component Study]{
        \includegraphics[width=0.48\linewidth]{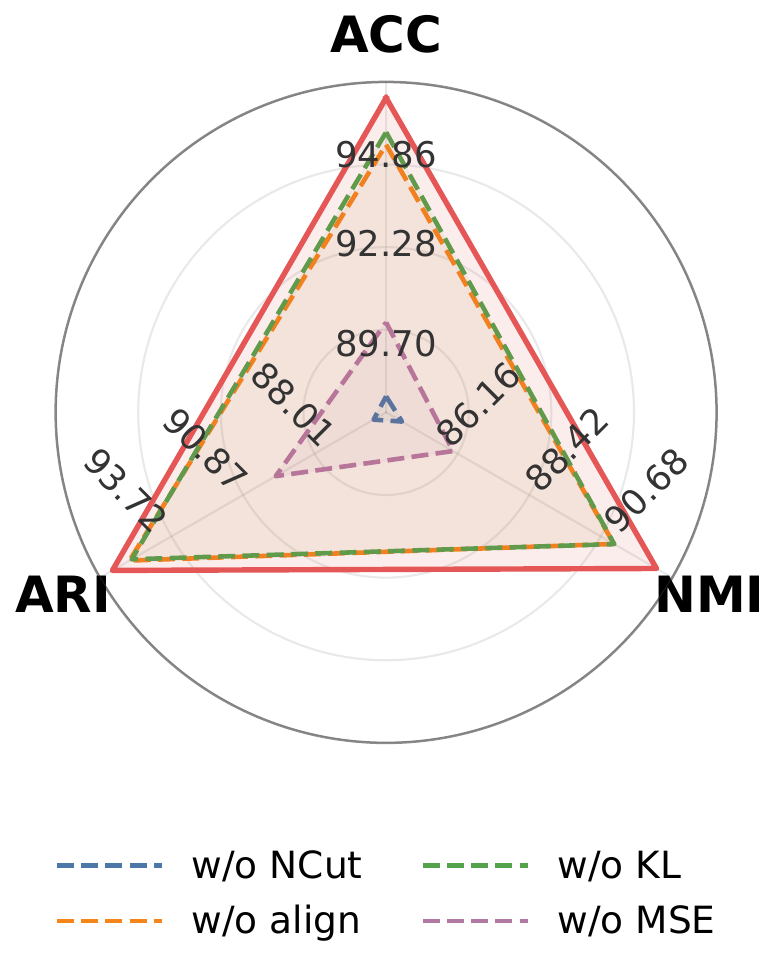}
        \label{fig:ablation_main}
    }
    \subfloat[Semantic Component Study]{
        \includegraphics[width=0.48\linewidth]{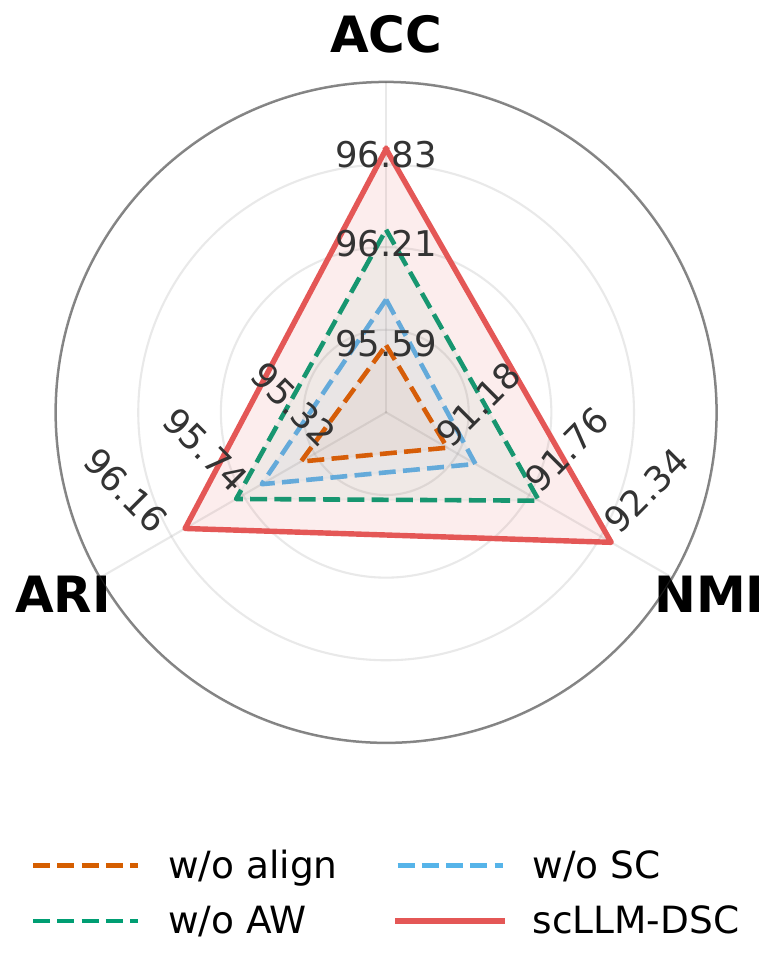}
        \label{fig:ablation_dual_semantic}
    }
    \caption{Ablation Study on \textit{Mauro Pancreas} Dataset.}
    \label{fig:ablation}
\end{figure}

\begin{table}[!t]
  \centering
  \resizebox{0.48\textwidth}{!}{%
  \begin{tabular}{l|ccc}
    \toprule
    \textbf{VARIANTS} & \textbf{ACC} & \textbf{NMI} & \textbf{ARI} \\
    \midrule
    Jasper-Token-Compression-600M & $96.27_{\pm1.12}$ & $91.63_{\pm1.42}$ & $95.70_{\pm0.83}$ \\
    granite-embedding-small-english-r2 & $96.36_{\pm0.90}$ & $91.53_{\pm1.51}$ & $95.71_{\pm0.69}$ \\
    Qwen3-Embedding-0.6B & $96.02_{\pm1.26}$ & $91.42_{\pm1.48}$ & $95.57_{\pm0.77}$ \\
    all-MiniLM-L6-v2 & $96.64_{\pm0.98}$ & $91.89_{\pm1.62}$ & $95.85_{\pm0.77}$ \\
    bge-m3 & $96.60_{\pm1.13}$ & $91.99_{\pm1.39}$ & $95.85_{\pm0.74}$ \\
    \textbf{OURS (text-embedding-3-small)} & $\textbf{96.94}_{\pm1.00}$ & $\textbf{92.43}_{\pm1.49}$ & $\textbf{96.08}_{\pm0.67}$ \\
    \bottomrule
  \end{tabular}}
  \caption{Robustness to LLM Variants on \textit{Mauro Pancreas}.}
  \label{tab:Ablation_LLM_variants}
\end{table}

\noindent\textbf{Framework Component Study.}
To evaluate the contribution of each module, we compare \methodname~with four variants on the \textit{Mauro Pancreas} dataset by systematically removing $\mathcal{L}_{align}$ (\textit{w/o align}), $\mathcal{L}_{NCut}$ (\textit{w/o NCut}), $\mathcal{L}_{MSE}$ (\textit{w/o MSE}), and $\mathcal{L}_{KL}$ ({w/o KL}).
Results visualized in Fig.~\ref{fig:ablation_main} reveal that~\methodname~consistently outperforms all ablation variants across all metrics, confirming that each component is vital for optimal performance. Notably, the significant performance degradation in \textit{w/o NCut} and \textit{w/o align} underscores the indispensable roles of both high-order structural manifolds and LLM-derived biological semantics. 
The performance gain demonstrates that \methodname~effectively synergizes high-order topological structures with explicit biological semantics, leading to more robust and interpretable cell clustering.


\noindent\textbf{Semantic Component Study.}
We evaluate the contribution of each semantic path by comparing \methodname~with three variants: \textit{w/o AW} (removing Abundance-Weighted Semantic Aggregation), \textit{w/o SC} (removing Sequence-based Contextual Encoding), and \textit{w/o align} (discarding both). 
Results in Fig.~\ref{fig:ablation_dual_semantic} show that while either path alone improves performance, their combination consistently yields the highest accuracy. 
This confirms that expression-weighted magnitudes and gene-order regulatory logic provide complementary biological insights into complex cellular states.


\subsection{Robustness to LLM Variants}
To assess the model's dependence on a specific language model, we investigate its robustness by substituting the default \textit{text-embedding-3-small} with various LLM backbones of differing architectures and scales. As reported in Tab.~\ref{tab:Ablation_LLM_variants}, \methodname~maintains remarkably stable performance across all evaluation metrics, with negligible fluctuations despite significant variations in the underlying embedding models. 
This stability indicates that \methodname~is highly compatible with diverse LLM architectures, as the specific backbone choice is not the primary determinant of performance. 
Combined with Fig.~\ref{fig:ablation}, these results underscore that integrating LLM-based semantics remains indispensable for superior accuracy, demonstrating the broad applicability of our framework.

\subsection{Parameter Sensitivity}
\begin{figure}[!t]
    \centering
    \includegraphics[width=0.55\linewidth]{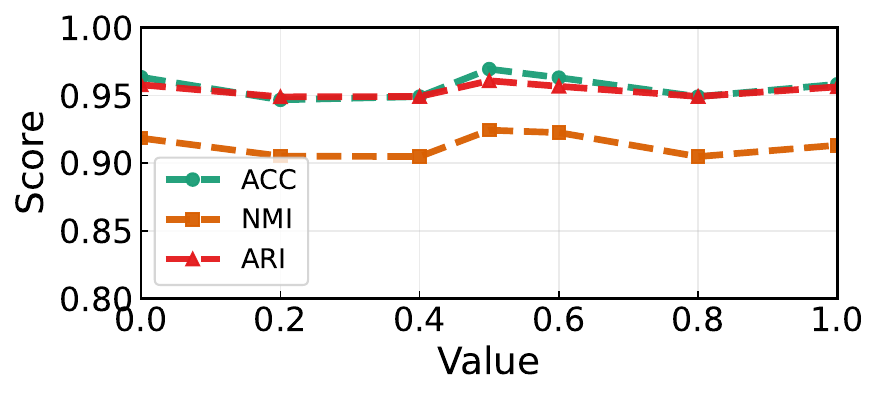}
    \caption{Parameter Sensitivity of $\omega$ on \textit{Mauro Pancreas dataset}.}
    \label{fig:omega_Parameter}
\end{figure}

We evaluate the impact of the balancing coefficient $\omega$ (Eq.~\ref{equ:cell_text_final}) on clustering performance. As illustrated in Fig.~\ref{fig:omega_Parameter}, metrics including ACC, NMI, and ARI fluctuate with varying fusion ratios of abundance and context semantics. Optimal results across all metrics are achieved at $\omega = 0.5$, confirming that the synergy of dual-path representations is essential for capturing comprehensive cell semantics. The performance decline at $\omega \in \{0, 1\}$ further proves that relying on a single semantic source is insufficient for maintaining peak clustering precision.

\section{Conclusion}
\methodname~presents a novel paradigm that integrates LLM-derived biological knowledge with deep structural clustering to address the persistent omission of semantic context in traditional scRNA-seq analysis.
By synergizing functional gene priors with graph-guided topological encoding, our framework establishes a unified latent space that aligns numerical transcriptomic features with high-level biological meaning. Benchmarks confirm that this task-specific cross-modal alignment not only achieves state-of-the-art accuracy but also provides an interpretable foundation for mechanism-aware discovery. Looking ahead, we plan to scale this framework to atlas-level datasets and incorporate spatial transcriptomics to further decode complex tissue architectures.

\clearpage
\section*{Acknowledgements}
This work is partially supported by the National Natural Science Foundation of China (Grant No. 92470204, 62406306, and 62406056) and the Guangdong Basic and Applied Basic Research Foundation (Grant No. 2024A1515140114).

\section*{Contribution Statement}
Ping Xu and Pengjiang Li contributed equally to this work. Specifically, they led the project, provided theoretical support, and were responsible for overall model design, code implementation, experimental design, and paper writing. 
Tian Du, Zaitian Wang, Jiawei Gu, and Zhiyuan Ning provided guidance in solving complex problems and assisted with paper writing. 
Ziyue Qiao, Pengfei Wang, and Yuanchun Zhou provided valuable feedback on manuscript drafts. All authors reviewed and approved the final version.



\bibliographystyle{named}
\bibliography{reference}

\end{document}